%% file: main.tex
\documentclass[a4paper]{svproc}
\pdfoutput=1
\usepackage[pdftex]{graphicx} %
\usepackage[dvipsnames]{xcolor}
\usepackage{cite}
\usepackage{multicol}
\usepackage[hidelinks]{hyperref}
\usepackage{bm}
\usepackage{rotating}
\usepackage{url}

\usepackage[mathscr]{euscript}  
\usepackage{amsmath} %
\usepackage{amssymb}  %
\usepackage{flushend}
\usepackage{multirow}\usepackage{multirow}

\usepackage{svg}
\usepackage{siunitx}
\usepackage[export]{adjustbox}
\usepackage{array}
\usepackage{booktabs}
\usepackage[ruled,vlined]{algorithm2e}  %

\usepackage[inline]{enumitem}
\usepackage[hang,flushmargin]{footmisc}
\usepackage{microtype}
\usepackage{threeparttable}
\usepackage{pifont}%
\usepackage{mathtools}

\usepackage{dsfont}

\usepackage{verbatim}

\definecolor{dark-green}{RGB}{12,80,12}

\usepackage{color, colortbl}
\definecolor{Gray}{gray}{0.9}
\usepackage{utfsym}
\newcommand{\yes}{\large \color{OliveGreen}\checkmark}
\newcommand{\no}{\color{BrickRed} \scalebox{1}{\usym{2613}}}

\usepackage{pgfplots}
\pgfplotsset{compat=1.16}

\newcommand{\website}{\url{https://semantic-search.cs.uni-freiburg.de}}

\DeclareSIUnit{\degree}{deg}
\DeclareSIUnit{\rad}{rad}
\pdfoutput=1

\newcommand{\secref}[1]{Sec.~\ref{#1}}
\renewcommand{\eqref}[1]{Eq.~(\ref{#1})}
\newcommand{\figref}[1]{Fig.~\ref{#1}}  
\newcommand{\tabref}[1]{Tab.~\ref{#1}}

\newcommand{\appref}[1]{Appx.~\ref{#1}}

\newcolumntype{P}[1]{>{\centering\arraybackslash}p{#1}}

\newcommand{\para}[1]{\parskip=5pt\noindent\textit{#1}}

\usepackage{tabularx}
\newcolumntype{Y}{>{\centering\arraybackslash}X}
\newcolumntype{Z}{>{\raggedleft\arraybackslash}X}

\let\llncssubparagraph\subparagraph
\let\subparagraph\paragraph
\usepackage[compact]{titlesec}
\let\subparagraph\llncssubparagraph

\begin{document}

\title{Perception Matters: Enhancing Embodied AI with Uncertainty-Aware Semantic Segmentation}
\titlerunning{Enhancing Embodied AI with Uncertainty-Aware Semantic Segmentation}

\author{Sai Prasanna$^1$\thanks{These authors contributed equally.}, %
        Daniel Honerkamp$^{1*}$, %
        Kshitij Sirohi$^{1*}$, %
        Tim Welschehold$^1$, %
        Wolfram Burgard$^2$, %
        and~Abhinav Valada$^1$%
        }

\institute{$^1$Department of Computer Science, University of Freiburg, Germany\\
$^2$Department of Engineering, University of Technology Nuremberg, Germany
}

\authorrunning{S. Prasanna~\textit{et al.}}
\maketitle

\let\thefootnote\relax\footnotetext{Funded by the Deutsche Forschungsgemeinschaft (DFG, German Research Foundation) - 417962828, the BrainLinks-BrainTools center of the University of Freiburg, and the Konrad Zuse School of Excellence in Learning and Intelligent Systems (ELIZA).}

\begin{abstract}
Embodied AI has made significant progress acting in unexplored environments. However, tasks such as object search have largely focused on efficient policy learning. %
In this work, we identify several gaps in current search methods: They largely focus on dated perception models, neglect temporal aggregation, and transfer from ground truth directly to noisy perception at test time, without accounting for the resulting overconfidence in the perceived state. 
We address the identified problems through calibrated perception probabilities and uncertainty across aggregation and found decisions, thereby adapting the models for sequential tasks. The resulting methods can be directly integrated with pretrained models across a wide family of existing search approaches at no additional training cost.
We perform extensive evaluations of aggregation methods across both different semantic perception models and policies, confirming the importance of calibrated uncertainties in both the aggregation and found decisions.
We make the code and trained models available at \website{}.
\end{abstract}

\section{Introduction}\label{sec:intro}
\input{1_introduction}

\vspace{0.2cm}
\section{Related Work}\label{sec:related}
\input{2_related_work}

\section{Technical Approach}\label{sec:approach}
\input{3_approach}

\section{Experimental Evaluations}\label{sec:experiments}
\input{4_experiments}

\section{Conclusion}\label{sec:conclusion}
\input{5_conclusion}

{\scriptsize
\bibliographystyle{spmpsci}
\bibliography{references.bib}}

\appendix

\input{6_appendix}
\end{document}

%% file: 1_introduction.tex
Embodied AI  has received a tremendous amount of attention in recent years, with new simulators enabling fast iteration in photorealistic, apartment-scale scenes based on real-world scans~\cite{szot2021habitat, kolve2017ai2}. This has been further advanced by institutionalized benchmarks and challenges in object search~\cite{habitatchallenge2023, robothor, embodiedAIworkshop}.
However, the main focus of this progress has primarily been on learning navigation and exploration policies. 
We analyze current object search works in \tabref{tab:related_work}, including, but not limited to, all top-performing approaches of the last three years' MultiOn, Habitat ObjectNav, and ImageNav challenges that released sufficient details.
As most of these methods require a large number of episode steps, the overwhelming majority of approaches train with ground truth perception and then evaluate with pretrained out-of-the-box semantic perception models. 
This zero-shot transfer to a learned model can significantly reduce training costs.
In return, we find it inducing a large gap between ground truth perception and semantic perception models, averaging over 25ppt, an error often as big as the gap to the optimal policy. 
Given their importance in the literature and the significance of the perception gap, this work focuses on the effective incorporation of such pretrained models.

\input{figures/barplot_comparison}

We identify a number of unsolved problems in the literature:
\begin{enumerate*}[label=(\roman*)]
    \item Many approaches have noted false detections as major failure source~\cite{zhou2023esc, chen2023not, luo2022stubborn}. However, due to their ease of use a vast share of the object search literature only evaluates comparatively old perception models. %
    \item Object search differs from pure perception tasks as it requires decision making over a sequence of observations. This time dimension is commonly ignored, opting to simply use the latest or most likely prediction without any aggregation over time~\cite{chaplot2020object, ye2021auxiliary, zhou2023esc}.
    \item Zero-shot transfer from a perfect ground truth perception to an imperfect perception model signifies that policies are unaware that they act upon imperfect perception, which is worsened by the fact that the pre-trained perception models are generally overconfident in their predictions~\cite{sensoy2018evidential}.
\end{enumerate*}

In order to quantify and address these problems, we first evaluate the impact of different semantic perception models and aggregation methods for sequential decision tasks. This differs from pure single-step perception evaluation based on the IoU (Intersection over Union) or precision~\cite{hurtado2022semantic}. In contrast, we measure the results over the full sequence of observations and actions, where early errors may impact or prevent later decisions.
\figref{fig:bar_comparisons} shows the large perception gap to ground truth semantics. While newer models can reduce this gap, we find that temporal aggregation on the perception level is a key to closing the gap.
To draw meaningful comparisons, we focus on one of the most used system structures, modular \textit{perception - mapping - policy} pipelines, in one of the most explored tasks, ObjectNav.
We introduce uncertainty-based aggregation and found decisions to address the identified problems. While previous methods develop complex, heuristic-based map aggregation strategies to cope with the overconfident and uncalibrated predicted probabilities~\cite{luo2022stubborn, staroverov2023skill}, we incorporate calibrated perception models with uncertainty estimation capabilities that can quantify this factor.
In the second step, we evaluate these models with a learned search policy and across different semantic models. We find that our conclusions generalize to different search strategies and semantic perception models.
The resulting methods directly integrate with existing approaches across a wide range of models without any additional training costs.

To summarize, this work makes the following main contributions:
\begin{itemize}[topsep=0pt]
    \item We identify and quantify gaps in the current integration of perception and aggregation methods in sequential embodied AI tasks.
    \item We incorporate calibrated perception probabilities and uncertainties in the agent's map aggregation and found decision making, thereby narrowing the gap to ground truth perception.
    \item We present extensive experimental evaluations and demonstrate that our approach can be easily integrated across a wide range of methods.
    \item We release the code at \website{}.
\end{itemize}

%% file: figures/barplot_comparison.tex
\pgfplotsset{select coords between index/.style 2 args={
    x filter/.code={
        \ifnum\coordindex<#1\def\pgfmathresult{}\fi
        \ifnum\coordindex>#2\def\pgfmathresult{}\fi
    }
}}

\pgfplotsset{%
    width=.5\linewidth,
    height=5.0cm
}
\pgfplotsset{%
    /pgfplots/ybar legend/.style={
        /pgfplots/legend image code/.code={
            \fill[##1] (0cm,0.6em) rectangle (0.9em,-0.3em);
}, },
}

\begin{figure}[t]
\centering
\scriptsize
\begin{tikzpicture}
\begin{axis}[
    ybar,
    axis lines*=left,
    legend style={at={(0.5,1.19)},
      anchor=north,legend columns=-1, draw=none},
    ybar legend,
    ylabel={Success Rate},
    symbolic x coords={gt,maskrcnn,segformer,emsanet},
    xtick=data,
    nodes near coords,
    nodes near coords align={vertical},
    enlarge x limits=0.25,
    ybar=4pt,
    ymin=0,
    ymax=80,
    bar width=15pt,
    xticklabels = {
            Mask-RCNN~\cite{he2017mask},
            Segformer \cite{xie2021segformer},
            EMSANet \cite{seichter2022efficient}
        },
  xticklabel style={
        text width={1.5cm},
        align={center},
      },
      legend style={/tikz/every even column/.append style={column sep=0.1cm}}]
    ]
\addplot[black,fill=blue!60] coordinates {(maskrcnn,34.7) (segformer,33.0) (emsanet,15.8)};
\addplot[black,fill=orange!100] coordinates {(maskrcnn,46.2) (segformer,54.3) (emsanet,56.4)};
\addplot[OliveGreen,sharp plot,update limits=false,line width=0.25mm] coordinates { ([normalized]0,75) ([normalized]3.5,75) };
\legend{One-Step Perception,Aggregated Perception,Ground-Truth Perception}
\end{axis}
\end{tikzpicture}
\vspace{-0.2cm}
\caption{Success rate of an RL agent~\cite{fabian22exploration} on the Habitat ObjectNav task with different semantic perception models. The gap from ground truth to learned perception models is often larger than the gap to an optimal policy. %
We propose uncertainty-based aggregation for sequential decision problems and find that this reduces the perception gap substantially.
Ground Truth: ground truth semantic masks, One-Step: latest semantic prediction, Aggregated: best evaluated aggregation method of the model (cf. \secref{sec:experiments}).}
\label{fig:bar_comparisons}
\vspace{-0.5cm}
\end{figure}
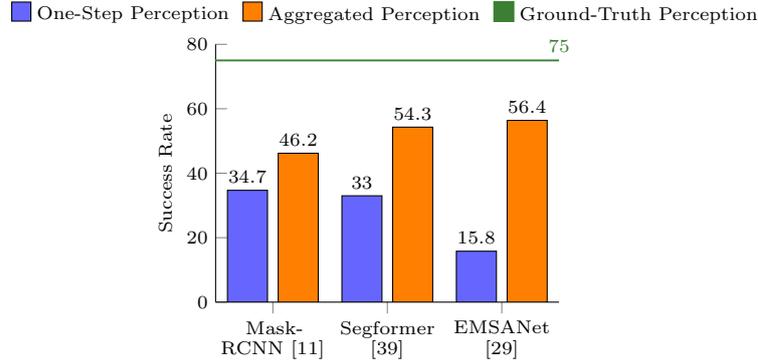

%% file: 2_related_work.tex
\input{tables/related_work}

\para{Object Search}
has been tackled by a wide range of methods, including classical approaches such as frontier exploration~\cite{yamauchi1997frontier}, vision-based reinforcement learning~\cite{chaplot2020learning}, or auditory signals~\cite{younes2023catch}. 
In recent years, map-based approaches that leverage semantic information have seen a large success~\cite{chaplot2020object, raychaudhuri2024mopa, zhu2022navigating, staroverov2023skill} across public challenges~\cite{embodiedAIworkshop}. On top of this map representation, both reinforcement learning (RL)~\cite{ramakrishnan2022poni, schmalstieg2023learning, chaplot2020object} and non-learned policies~\cite{zhou2023esc, luo2022stubborn} have been used successfully. This modularized network structure has found use in tasks beyond ObjectNav~\cite{chang2023goat, honerkamp2024language}.
We provide a comprehensive overview of recent approaches, including the top performing approaches of the last three year's ObjectNav~\cite{habitatchallenge2023}, ImageNav~\cite{habitatchallenge2023} and MultiOn~\cite{wani2020multion} challenges, in \tabref{tab:related_work}. We make a number of observations: 
\begin{enumerate*}[label=(\roman*)]
    \item Due to the large size of perception models, learning-based approaches opt to use ground truth perception during training, to then replace this module with a pretrained perception model at test time. Occasionally, the perception model is first finetuned on the dataset~\cite{maksymets2021thda}.
    \item Due to their ease of use, the majority of approaches rely on comparably old perception models such as Mask-RCNN~\cite{he2017mask} or Rednet~\cite{jiang2018rednet}. Only a few works are able to stem the work to integrate recent state-of-the-art models~\cite{staroverov2023skill, zemskova2024interactive, zhou2023esc}.
    \item While the policy networks often integrate previous information through RNNs, the time dimension of these sequential decision making processes is largely ignored for the perception level, and simply the last prediction is used.
    While separated works integrate aggregation methods~\cite{luo2022stubborn, staroverov2023skill}, comprehensive comparisons within sequential decision tasks are missing. Outside this literature, methods employ techniques such as Bayesian update and probability averaging for map aggregation utilizing the non-calibrated probabilities from the perception semantic perception models~\cite{thrun2002probabilistic, mccormac2017semanticfusion, zurn2022trackletmapper,gosala2023skyeye}. 
    \item As a result, we find a large test-time gap between ground truth semantics and deployment with a learned model, averaging 25.8 ppt, with false detections often identified as the largest source of errors~\cite{chen2023not}.
\end{enumerate*}

\para{Uncertainty-Aware Perception}: Uncertainty estimation techniques for perception can be classified into sampling-based and sampling-free methods. Sampling-based methods require either multiple forward passes~\cite{gal2016dropout} or multiple networks~\cite{lakshminarayanan2017simple} to estimate the uncertainty. Therefore, they require significant time and memory and, thus, are not suitable for real-time applications. Hence, sampling-free methods have gained more attention in recent research. Sensoy~\textit{et~al.}~\cite{sensoy2018evidential} proposed evidential learning for the classification task to learn parameters of a high-order distribution from which the classification uncertainty can be inferred. Other works have adapted evidential learning for object detection~\cite{nallapareddy2023evcenternet} and segmentation tasks with different modalities~\cite{sirohi2023uncertainty,sirohi2023uncertainty2}. Sirohi~\textit{et~al.}~\cite{sirohi2024uplam} further utilized the evidential output from~\cite{sirohi2023uncertainty} and proposed the methods for uncertainty-aware mapping. However, methods relying on evidential learning need to be trained from scratch. It is possible to directly utilize the softmax probabilities from the segmentation network in probabilistic log-odds softmax mapping~\cite{zurn2022trackletmapper}. However, the resulting map also creates over-confident uncertainty estimations~\cite{sirohi2024uplam}. 
Guo~\textit{et~al.}~\cite{guo2017calibration} proposed temperature scaling to calibrate the softmax probabilities obtained from the network by scaling the predicted output logits on the validation set. This enables the use of any pre-trained network by tuning the scaling factor without the need for costly re-training. %
Hence, we utilize temperature scaling to obtain calibrated probabilities and uncertainties for semantic segmentation models.

%% file: tables/related_work.tex
\begin{table}
    \centering
    \vspace{-0.5cm}
    \fontsize{6pt}{6pt}\selectfont
    \caption{Comparison of existing object search approaches.}
    \label{tab:related_work}
    \begin{threeparttable}
    \setlength{\tabcolsep}{1pt}
    \begin{tabularx}{\textwidth}{lcccYccc}
      \toprule
        Paper & Train & Test & Perception & Found & \multicolumn{3}{c}{Success Rate}\\
              & Perception & Perception & Aggregation  & Decision & GT & Model & Gap\\
      \midrule
        SemExp~\cite{chaplot2020object} & Mask-RCNN (PT)& Mask-RCNN (PT)& MaxPool & Distance & 73.1$^1$ & 54.4$^1$ & 18.7 \\
        EEAux~\cite{ye2021auxiliary} & GT & RedNeT (FT) & -- & RL & 58.0$^{2c}$ & 34.6$^2$ & 23.4\\
        PONI~\cite{ramakrishnan2022poni} & GT & Mask-RCNN (FT) & Max-Pool & Distance & 86.5$^1$ & 73.6$^1$ & 12.9\\
        PONI~\cite{ramakrishnan2022poni} & GT & RedNeT (FT) & Max-Pool & Distance & 58.2$^3$ & 31.8$^3$  & 26.4\\
        Zhu~et~al.\cite{zhu2022navigating} & GT & Mask-RCNN (PT) & -- & Distance & 76.8$^8$ & 43.8$^8$ & 33.0\\
        Stubborn~\cite{luo2022stubborn} & -- & RedNeT (FT) & Agg. Scores & Naive Bayes & 67.0$^{2a}$ & 37.0$^{2b}$ & 30.0\\
        Schmalstieg~\cite{fabian22exploration} & GT & GT & -- & GT & 85.7$^9$ & -- & --\\
        HIMOS~\cite{schmalstieg2023learning} & GT & GT & -- & GT & 95.6$^{9b}$ & -- & --\\
        MOPA~\cite{raychaudhuri2024mopa} & -- & Detic (PT) & No Forgetting & RL (PT) & 81.0$^7$ & 29.0$^7$ & 52.0\\
        PIRLNav~\cite{ramrakhya2023pirlnav} & Demos & E2E & -- & RL & 61.9$^6$ & -- & --\\
        ESC~\cite{zhou2023esc} & -- & GLIP (PT) & -- & Distance & 64.0$^{6a}$ & 39.2$^6$ & 24.8   \\
        Dragon~\cite{chen2023not} & -- & Mask-RCNN (PT) & Visual SLAM$^\dagger$ & Distance & 84.2$^1$ & 57.6$^1$ & 26.6\\
        SkillFusion~\cite{staroverov2023skill} & GT, part learned & SegFormer$^*$& Binary w/ decay & RL w/ Not-sure-action & 64.7$^5$ & 54.7$^5$ & 10.0\\
        SkillTron~\cite{zemskova2024interactive} & GT, part learned & SegmATRon & Binary w/ decay & Distance & -- & 59.0$^4$ & -- \\
        \rowcolor{Gray}
        \textbf{Average} &&&&&&& \textbf{25.8} \\
      \bottomrule
    \end{tabularx}
        \begin{tablenotes}[para,flushleft]
          \fontsize{6pt}{6pt}\selectfont
           GT: ground-truth information. 
           PT: pretrained, 
           FT: finetuned. 
           Distance: if target is mapped, navigate to it, stop when close enough. 
           No Forgetting: detected objects will never be deleted. 
           Aggregated Scores: number of views, sum and max of class predictions. 
           Demos: behavior cloning on expert demonstrations. 
           Max-Pool: Channel-wise max-pooling of current and previous map. 
           Part learned: GoalReacher trained with 20\% noisy semantics. 
           Not-sure action: the agent can decide to return to exploration if not confident to terminate the episode. 
           Detection w/ decay: Accumulation of binary detections with decay coefficient.
           E2E: end-to-end learned, without explicit mapping.
           $^*$no details provided if pretrained or trained from scratch.
           $^\dagger$no details on fusion provided.\\
           Note that absolute success rates are not comparable across tasks.
           $^1$Gibson dataset, SemExp split.
           $^2$MP3D, Habitat Challenge 2020 task, validation set, with 6$^a$ or 15$^b$ out of 21 target objects, $^c$300 episode subset.
           $^3$MP3D, Habitat Challenge 2021 task, validation set.
           $^4$HM3D v0.2, Habitat Challenge 2023 task, test set.
           $^5$HM3D, 20 scenes of validation set.
           $^6$HM3D, Habitat Challenge 2022, validation set, $^a$Ground-truth SR estimated from detection error rate.
           $^7$HM3D, MultiOn 2.0.
           $^8$MP3D with 6 target categories and 10 test scenes (split unspecified).
           $^9$iGibson Challenge 2020, test set, $^b$interactive search.
         \end{tablenotes}
    \end{threeparttable}
    \vspace{-0.75cm}
\end{table}

%% file: 3_approach.tex
\subsection{Problem Statement}
In ObjectNav, the agent starts in an unexplored environment and has to find and navigate to an instance of a target category $c$, using only its RGB-D camera and localization. We follow the definition of the Habitat 2023 ObjectNav Challenge~\cite{habitatchallenge2023} in the HM3D dataset~\cite{ramakrishnan2021hm3d}. An episode is considered successful, if the agent issues a \texttt{Stop} action within \SI{1}{\meter} of the target object and the object can be viewed by an oracle from the stopping location. The episode terminates unsuccessful if no found decision is made by 1,000 steps.
We use the continuous action parameterization with some adaptations: %
\begin{enumerate*}[label=(\roman*)]
    \item We flip the camera resolution from vertical to a more common landscape resolution of $640\times 480$ and set the camera pitch to \SI{-20}{\degree}.
    \item We use only those target classes that are covered by all pretrained perception models, %
    i.e. we omit the \emph{plant} class.
    \item We adapt the step-size for continuous actions from \SI{0.1}{\meter\per\second} to \SI{1.0}{\meter\per\second}, with a maximum linear velocity of \SI{0.25}{\meter\per\second}. This ensures the agent can travel the same maximum distance in the allowed time budget as in the discrete action parameterization. %
    \item We fix a bug that resulted in no collisions detection and low-level velocity integration to take multiple steps for larger velocities. 
\end{enumerate*}

\input{figures/pipeline}

\subsection{Model Structure}
In this work, we focus on modular approaches that have seen widespread use and achieved state-of-the-art results~\cite{chaplot2020object, ramakrishnan2022poni, zhu2022navigating, luo2022stubborn, fabian22exploration, schmalstieg2023learning, raychaudhuri2024mopa, zhou2023esc, chen2023not, staroverov2023skill, zemskova2024interactive}, depicted in \figref{fig:pipeline}.
These approaches first obtain semantic information for the current camera image. They then use a mapping module to obtain an explicit representation, by projecting this data into a top-down map and integrating it into a global map. Finally, an agent uses this representation to make navigation and found decisions - commonly via an ego-centric map for reinforcement learning agents, or using the full global map for planners.
The modularity of this pipeline enables the use of different segmentation models with different policies. We incorporate uncertainty and calibrated probabilities into these systems and leverage them for efficient temporal aggregation and found decisions.

\subsection{Uncertainty-Aware Perception}

We assume access to a pretrained semantic segmentation network which takes RGB and/or depth images as input and predicts a semantic segmentation output as a logits vector $l= [l^{1}, ..., l^{C}]$ over the pixels, where $C$ is the number of classes. 
However, these semantic segmentation models are commonly trained with a cross-entropy loss~\cite{xie2021segformer, seichter2022efficient, jiang2018rednet} which includes employing a softmax operation on the logits, which inflates the predicted probability of the class. Thus, we employ the temperature scaling technique as proposed by~\cite{guo2017calibration}, where they tune a scaling factor on the validation set to scale the logits for better probability estimates. However, as the networks are pre-trained on different datasets, we tune the scaling factor $t$ on a labeled set from the HM3D dataset~\cite{ramakrishnan2021hm3d}. Please note that this image set is independent of any policy training.
Finally, we utilize the scaled logit vector $l_s= [\frac{l}{t}^{1}, ..., \frac{l}{t}^{C}]$ to obtain the probability vector $p_i = \text{softmax}(l_s)$. Where, $p_i^{pred}= [p_i^{1}, ..., p_i^{C}]$ consists of per-class classification probability for the pixel $i$ in the image. Finally, we employ normalized entropy to calculate the corresponding per-pixel classification uncertainty $u_i \in {[0,1]}$, defined by $ {u}_i =  \frac{\sum_{c = 1}^C p^c_i \log(p^c_i)}{\log(C)}$.

\subsection{Perception Uncertainty Weighted Map Aggregation}
\label{sec:map_agg}
We use a Birds-Eye-View (BEV) grid map representation with a grid cell size of 3$\times$3 cm$^2$. We project the perception prediction to the respective cell based on the depth image and the ground truth pose of the robot, using the top most voxel. %
We utilize the calibrated probability vector $p^{pred}$, together with the perception uncertainty $u$ for the mapping. For every grid cell $k$ in the map and $N$ measurements, we calculate a weighted average of the probability vector, weighted by the inverse of the perception uncertainty, to obtain the aggregated probability vector $p_k = \frac{1}{U} \sum_{n=1}^N \frac{1}{u_{n,k}} p^{pred,k}$ of size $C\times1$, where $U = \sum_{n=1}^N \frac{1}{u_{n,k}} $. Similar to the perception, we calculate the mapping uncertainty $u_k^{map}$ for every grid cell as the normalized entropy obtained from the probability vector $p_k$. Thus, we maintain a vector $m_k$ of size $C+3$ for every cell $k$ of the map, where $m_k$ = [$p_k$, $height$, $occupancy$, $u_k^{map}$]. The $height$ keeps track of the maximum encountered height within a cell, and occupancy is a binary value, which we set to one if the $height > \SI{0.1}{\meter}$, and zero otherwise.

\subsection{Map Uncertainty-based Found Decision}
We issue a found decision when the target object follows two criteria, \begin{enumerate*}[label=(\roman*)] \item the target object is within \SI{1}{\meter} of the robot, and
\item the map uncertainty, $u_k^{map}$, of the map cell occupied by the target object is less than the threshold $\xi = 0.4$. We set the value of the threshold empirically based on hyperparameter optimization on a training set (cf. \secref{sec:baselines}).
\end{enumerate*}
The idea is that perception is prone to make errors at farther distances. Hence, the distance constraint ensures the robot is in the vicinity of the target object when the decision is made. The map uncertainty is crucial to filter false positives caused by varying perception predictions. If the perception shows variation in the predictions, the uncertainty will be high. Thus, we only mark an object as found when perception provides multiple observations with low uncertainty for the target object.

\subsection{Policies}\label{sec:policy}
While the perception literature has developed reliable metrics for single image evaluation,
we are interested in the performance of perception models in sequential decision making problems. As such, the performance of perception and policy are tightly interwoven as the perceived state will impact the policy's next actions which in turn will alter the next state. As a result, some errors may have a larger impact than others, if they lead the agent astray early on or terminate the episode too soon. For this reason, we evaluate two settings:

First, we propose the use of a state-independent ground truth shortest path policy to evaluate all methods on the same sequence of observations to isolate the impact of the perception and aggregation components.
We leverage Habitat's shortest path implementation~\cite{szot2021habitat} in the ground truth navigation mesh with the target object as the goal. For comparability, we collect metrics over the full trajectory until the goal, even if a false found decision would terminate the episode early.

In the second step, we evaluate how these results transfer to different policies and impact the overall performance within the decision making loop.
For this, we implement a recent reinforcement learning based object search policy~\cite{fabian22exploration} that follows the modular model structure shown in \figref{fig:pipeline}. Given the target class, the agent transforms the full semantic map into a local and global ego-centric map and maps the task-relevant objects to a \textit{target-object} color and non-relevant objects to an \textit{occupied} color.
Its policy network then predicts both the most likely direction toward the target object and navigation commands.
The agent is trained with PPO with the ground truth semantic perception and then deployed with the learned perception model and aggregation strategy.

%% file: figures/pipeline.tex
\begin{figure}
  \centering
  \scriptsize
  \vspace{-0.5cm}
  \includegraphics[width=\textwidth]{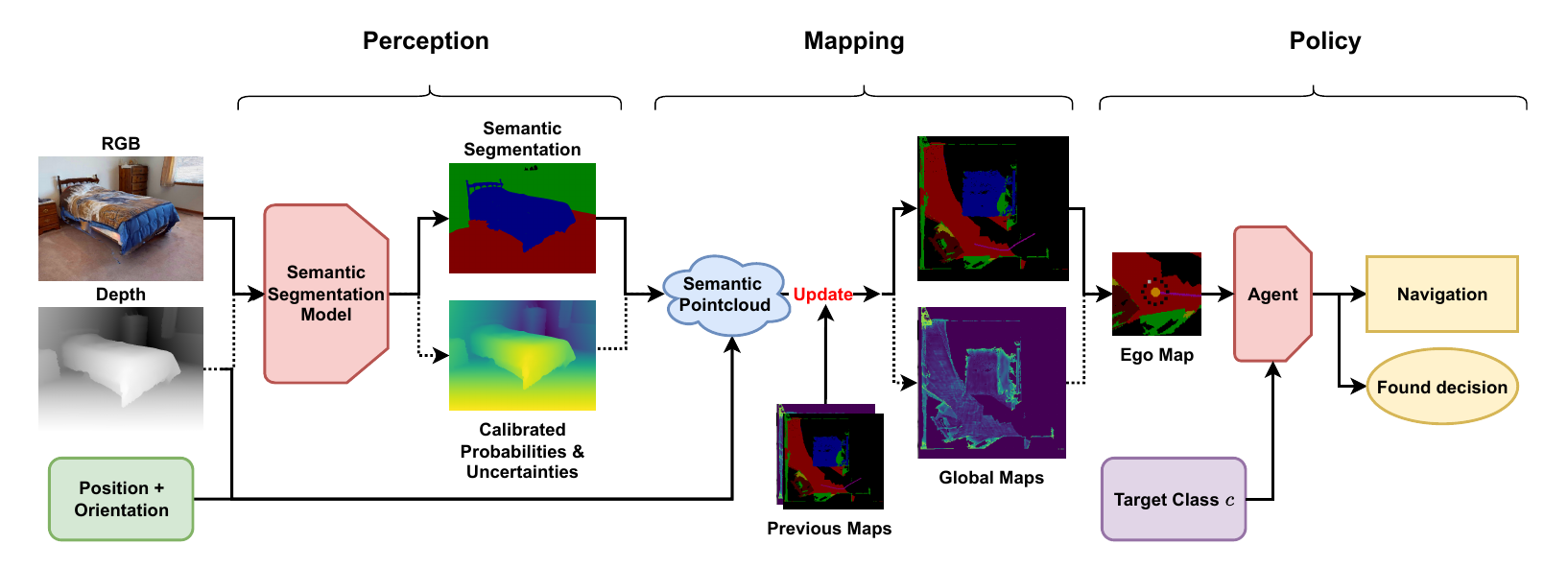}
  \caption{Overview of modular object search pipelines. First, a semantic segmentation model classifies the current image. A mapping module then fuses this information into a semantic point cloud and integrates it into a global map. From this map, either an egocentric map is extracted for RL agents or the full map is used by a planner. An agent then determines navigation and found decision for a given target class $c$. We develop general methods to incorporate calibrated uncertainties in this system for temporal aggregation of the semantic perception and consistent found decisions.} \label{fig:pipeline}
\vspace{-0.5cm}
\end{figure}

%% file: 4_experiments.tex
We evaluate our approach across a wide range of semantic perception models and multiple policies on the validation split of the Habitat 2023 ObjectNav Challenge~\cite{habitatchallenge2023} in the HM3D dataset\cite{ramakrishnan2021hm3d}.
    
\noindent{\textbf{Metrics}}: We compute the following evaluation metrics:\\
\para{Success Rate (SR):} The share of successful episodes in which the agent found the target and correctly raised the found decision within the time limit.\\
\para{Found/False Positive Rate (FPR):} Share of episodes with incorrect found decision.\\
\para{Found/False Negative Rate (FNR):} Share of episodes in which the agent failed to raise a found decision.\\
\para{Detection/False Positives (\#FP):} Number of times the target object was incorrectly mapped and shown to the agent per episode. Note that multiple false detections can occur over a single episode. To count the number of object detections (as opposed to pixels), we dilate all target class predictions outside the ground truth bounding box on the map and count each connected component as a false detection.\\
\para{Detection/False Negatives (\#FN):} Number of times per episode a target object existed in the map created with ground truth semantic camera, but no target class was shown to the agent within the ground truth bounding box of that object.\\
\para{Success weighted by Path Length (SPL):} Success weighted inverse normalized length of the agent's path.

\noindent{\textbf{Baselines}}\label{sec:baselines}:
We compare against a large range of baselines for aggregation and found decision: \\
\para{Ground Truth:} Build the map with the ground truth semantic camera images from the simulator. Always trigger found decision if close enough to mapped target.\looseness=-1\\
\para{Latest:} Map the class with the highest predicted probability at each step, overwriting any previous values. Found decision is made if the agent is within the success radius (\SI{1}{\meter}) of the grid cell containing the goal object.\\
\para{Hits/Views:} Based on the probabilistic counting method~\cite{thrun2002probabilistic} for aggregating multiple observations for binary classifications and views: Maintain additional map channels to track the number of hits (detections) of the goal class for each cell and how often this cell was in close view from a distance below $d_{view}$. If the agent is within the distance threshold of \SI{1}{\meter} of a cell with at least $v$ views and the hits/view ratio is above $\theta$, raise the found decision. Otherwise, classify as false detection and stop mapping the target in that cell.\\
\para{Skillfusion~\cite{staroverov2023skill}:} Erode the goal object in the local map with a kernel of $\SI{4}{\centi\meter}\times \SI{4}{\centi\meter}$ to remove outliers. Then maintain a grid map with continuous values representing the existence of goals. This value is incremented by one if a goal object gets projected to the cell. Otherwise, the grid cell value is multiplied by a decay factor $\alpha$. The goal is mapped for the agent if the cell value is greater than a threshold $T$. If the agent is within found distance of such a grid cell, raise the found decision.\\
\para{Stubborn~\cite{luo2022stubborn}:} Maintain additional map channels with total views, cumulative confidence, maximum confidence and maximum non-target confidence. These features are given to a Naive Bayes Classifier that outputs a binary found decision. As the trained classifier is not released, we train it on features collected from 64 episodes collected with the shortest path policy in unique training scenes.\\
\para{Latest Filtered:} Same as Latest, but only map the target category if the map uncertainty for the mapped target object is below a threshold $\rho$. Otherwise, map as \textit{occupied} class.\\
\para{Log Odds:} Bayesian updating of the grid cells with a multi-class log odds vector for each cell~\cite{zurn2022trackletmapper} but using our calibrated probabilities. The agent is shown the most likely class at each step.
If a goal object is within found distance and the uncertainty of the posterior is less than a threshold $\xi$, mark the object as found.

We set the values of the parameters for all methods via hyperparameter search with a tree-structured parzen estimator \cite{bergstra2011algorithms}. %
For each method, we set a budget of 20 trials and evaluate the value of each configuration as the average success rate over 30 episodes of the training scenes collected with the shortest path policy, and found the optimization process to converge to stable parameter values.

\noindent{\textbf{Perception Models}: To evaluate the generalizability of the aggregation methods, we evaluate all approaches across different semantic models.\\
\para{Mask-RCNN~\cite{he2017mask}:} The most used perception model in the ObjectNav literature (cf. \tabref{tab:related_work}), representative of simple out-of-the-box models. Pretrained on MS-COCO. As Mask-RCNN is an instance segmentation model, it does not provide class probability for all pixels, hence not all aggregation methods are applicable.\\
\para{Segformer~\cite{xie2021segformer}:} Transformer-based base model used by most recent ObjectNav challenge winners~\cite{staroverov2023skill}. It is pretrained on ADE20k \cite{Zhou_2017_CVPR}.\\
\para{EMSANet~\cite{seichter2022efficient}:} A recent panoptic segmentation model which takes RGB and depth images as input. We only utilize semantic segmentation prediction, i.e., we only reason about the class of individual pixels and do not use the instance segmentation results of EMSANet. The model is pre-trained on the SUN RGB-D~\cite{song2015sun} and Hypersim~\cite{roberts2021hypersim} datasets.

\input{figures/ece.tex}
\subsection{Evaluation of Uncertainty Estimation}

We evaluate the probability and uncertainty estimation quality of the perception models through the Expected Calibration Error (ECE)~\cite{naeini2015obtaining} and Uncertainty Expected Calibration Error (uECE)~\cite{sirohi2023uncertainty}. The calibration of a network defines how well the predicted confidence matches the actual accuracy of the prediction. ECE quantifies the error between the maximum class probability and the accuracy, whereas uECE measures the error between the network confidence and the accuracy, where confidence is defined as $1 - \text{uncertainty}$. As the map aggregation as defined in the (\secref{sec:map_agg}) uses both uncertainty and probability, it is desirable to have both ECE and uECE to be low for better performance.

In~\figref{fig:ece}, we show the calibration plots for vanilla and temperature-scaled versions of Segformer and EMSANet. A perfect calibration corresponds to the solid black line in the plot. As we can see, both ECE and uECE decrease after applying temperature scaling for both networks. While the ECE for EMSANet with temperature scaling (TS) decreased from 26 percent to 5 percent, Segformer's only decreased by one percentage point. Thus the probabilities become much better calibrated with temperature scaling for EMSANet than for Segformer. Hence, we expect better average probability aggregation for EMSANet in comparison to Segformer. However, the uncertainty calibration of both EMSANet and Segformer shows a significant improvement in uECE of 17 and 10 percent respectively. Thus, uncertainty weighting should have an identical impact for both Segformer and EMSANet. However, as EMSANet has both better calibrated probability and uncertainty, we find it to be better for the object search task.

\subsection{Perception's Impact on Sequential Navigation}
\input{tables/emsanet_sp}
To isolate the impact of the perception in sequential navigation tasks, we first compare the perception-based aggregation strategies over identical sequences based on the shortest path policy. The results for the best performing semantic model, EMSANet, are reported in \tabref{tab:emsanet_sp}. 
We find that the shortest path policy achieves nearly perfect performance with the oracle ground truth perception, signifying any drop in the success rate can be attributed to the perception gap. Note that we find a very small number of six unsolvable episodes due to missing or incorrect targets.
We find that temporal aggregation is essential, as simply using the latest predictions results in a high number of false found decisions, leading to a significant drop to 30\%, while aggregation methods across the board reduce this gap. %
Secondly, we find both calibrated probabilities and uncertainty-based found decisions essential, outperforming the heuristic-based methods of using counts or erosion to eliminate false positives. Finally, we find our proposed averaging to provide the most reliable update. 
Qualitative examples are shown in \figref{fig:map_viz}. We can see that the agent is able to filter out the false positives (shown in circles), whenever the underlying map uncertainty is high for those objects.

\input{figures/map_viz}

\subsection{Perception's Impact on Sequential Decision Making}
\input{tables/emsanet_rl}
We then investigate the interdependence between the aggregation methods and policy, and evaluate the methods with the learned RL agent. The agent is trained with ground truth perception and deployed with the learned perception as described in~\secref{sec:policy}. Results are presented in \tabref{tab:emsanet_rl}.
The agent achieves a success rate of 75\% with ground truth perception, indicating a policy gap of 25\%. Across aggregation methods, we find that our results generalize to different policies, confirming the previous conclusions on the shortest path policy. Calibrated perception probabilities and uncertainties are essential which with our proposed aggregation method achieves a significant reduction in the perception gap, both in terms of success rate and path efficiency as measured by the SPL.

\subsection{Generalizability Across Perception Models}
\figref{fig:bar_comparisons} compares the results across the different perception models. We find the importance of calibrated aggregation confirmed across the board. While Mask-RCNN's object segmentation performs well in single image prediction, using newer models can close the often reported perception gap by over 10ppt with accurate aggregation. 
The full results across all aggregation methods, reported in \appref{app:aggregation_results}, confirm the relative results across aggregation methods, as the relative ranking of methods remains very stable. We find the proposed weighted averaging to consistently perform best or on par across models and policies. One exception is the Skillfusion aggregation in combination with Segformer which it was developed with, evaluated with the RL policy. While achieving the same SPL, it reaches a 1.4ppt higher success rate in this setting. However, it is expected as we can see from the results in \figref{fig:ece}, that even after temperature scaling, the Segformer model's probabilities do not calibrate well. Nonetheless, utilizing calibrates uncertainties helps our method achieve a performance close to Skillfusion. This result further bolsters our claim that better calibrated networks help in better navigation tasks.

%% file: figures/ece.tex
\begin{figure}[t]
\centering
\includegraphics[width=0.45\textwidth,trim={0.5cm 1.5cm 0.25cm 1.0cm}]{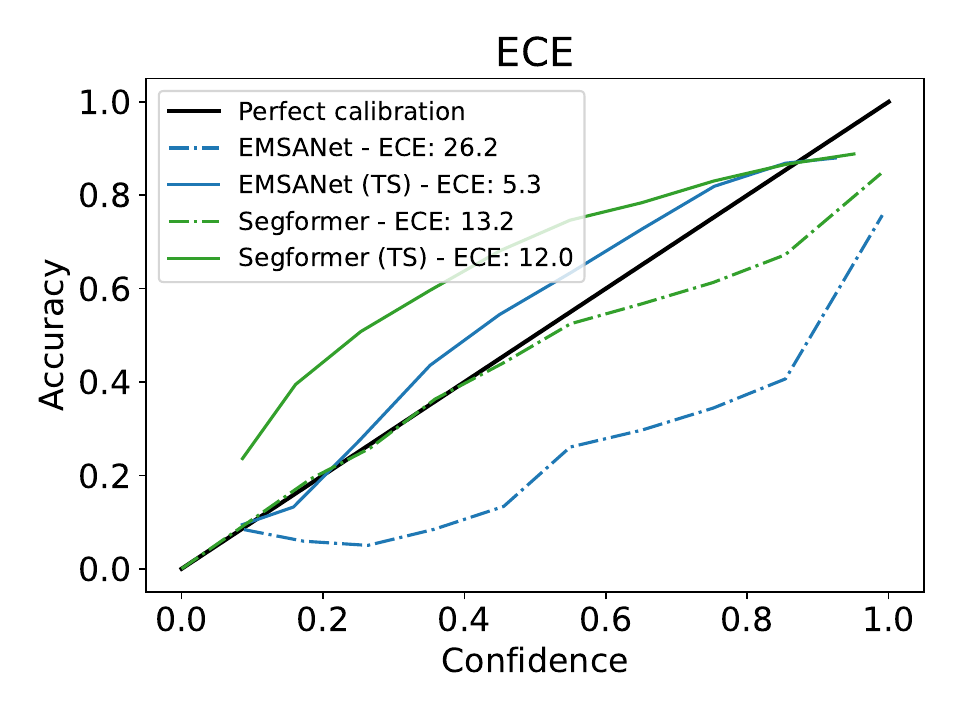}
\includegraphics[width=0.45\textwidth,trim={0.5cm 1.5cm 0.25cm 1.0cm}]{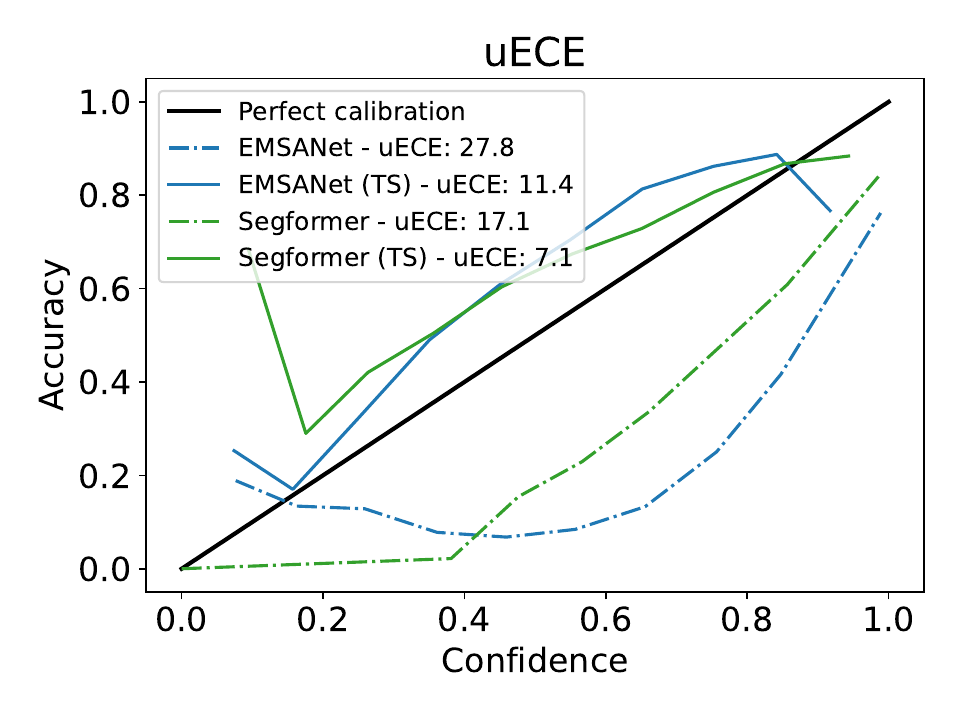}
\caption{Expected Calibration Error (left) and Uncertainty Expected Calibration Error  (right) of the different semantic perception models on the validation set.}
\label{fig:ece}
\vspace{-0.3cm}
\end{figure}

%% file: tables/emsanet_sp.tex
\begin{table*}[ht]
\scriptsize
\centering
\caption{ObjectNav results of the Shortest-Path Policy with EMSANet.}\label{tab:emsanet_sp}
\begin{threeparttable}
    \begin{tabularx}{\textwidth}{l|c|c|YYY|YY}
        \toprule
        & \textbf{Calibrated} & \textbf{Uncertainty} &  \textbf{SR}  & \multicolumn{2}{c|}{\textbf{Found}} & \multicolumn{2}{c}{\textbf{Detection}}\\
        \textbf{Aggregation} & \textbf{Probabilities} & \textbf{Found Decision} &                                                          &   FPR & FNR & \#FP & \#FN \\
        \midrule
        \rowcolor{Gray}
        Ground Truth & \no & \no  & 99.2    & \phantom{0}0.0      & \phantom{0}0.8      & 0.11           & 0.00           \\
        Latest & \no & \no & 30.1 & 67.0 & \phantom{0}3.0 & 5.81 & 0.08 \\
        Hits/Views & \no & \no & 69.2 & 12.4 & 18.4 & 0.52 & 0.19 \\
        Skill Fusion~\cite{staroverov2023skill} & \no & \no & 67.8 & 20.3 & 11.9 & 0.73 & 0.26 \\
        Stubborn~\cite{luo2022stubborn} & \no & \no & 32.7 & 63.8 & \phantom{0}3.5  & 5.49 & 0.08  \\
        Latest Filtered & \yes & \no & 71.7 & \phantom{0}7.9 & 20.4 & 0.28 & 0.46 \\
        Log odds   & \yes & \yes & 70.3 & 19.1 & 10.6 & 4.52 & 0.08 \\
        Averaging  & \yes & \no & 44.7 & 52.0 & \phantom{0}3.3 & 4.34 & 0.08 \\
        Averaging & \no & \yes &  47.9 &  48.2 &  \phantom{0}3.9 &  4.48 & 0.08 \\
        Averaging (Ours) & \yes & \yes & \underline{73.8} & \phantom{0}8.5 & 17.7 & 4.57 & 0.08 \\
        Weighted Averaging (Ours)  & \yes & \yes & \textbf{74.9} & \phantom{0}8.7 & 16.4 & 4.58 & 0.08 \\
        \bottomrule
    \end{tabularx}
      \begin{tablenotes}[para,flushleft]
       Best and second best in bold and underlined. SR: success rate, FPR: false positive rate, NR: false positive rate, \#FP: false positives, \#FN: false negatives.
     \end{tablenotes}
 \end{threeparttable}
 \vspace{-0.5cm}
\end{table*}

%% file: figures/map_viz.tex
\begin{figure}[t]
  \centering
  \scriptsize
  \includegraphics[width=0.8\textwidth]{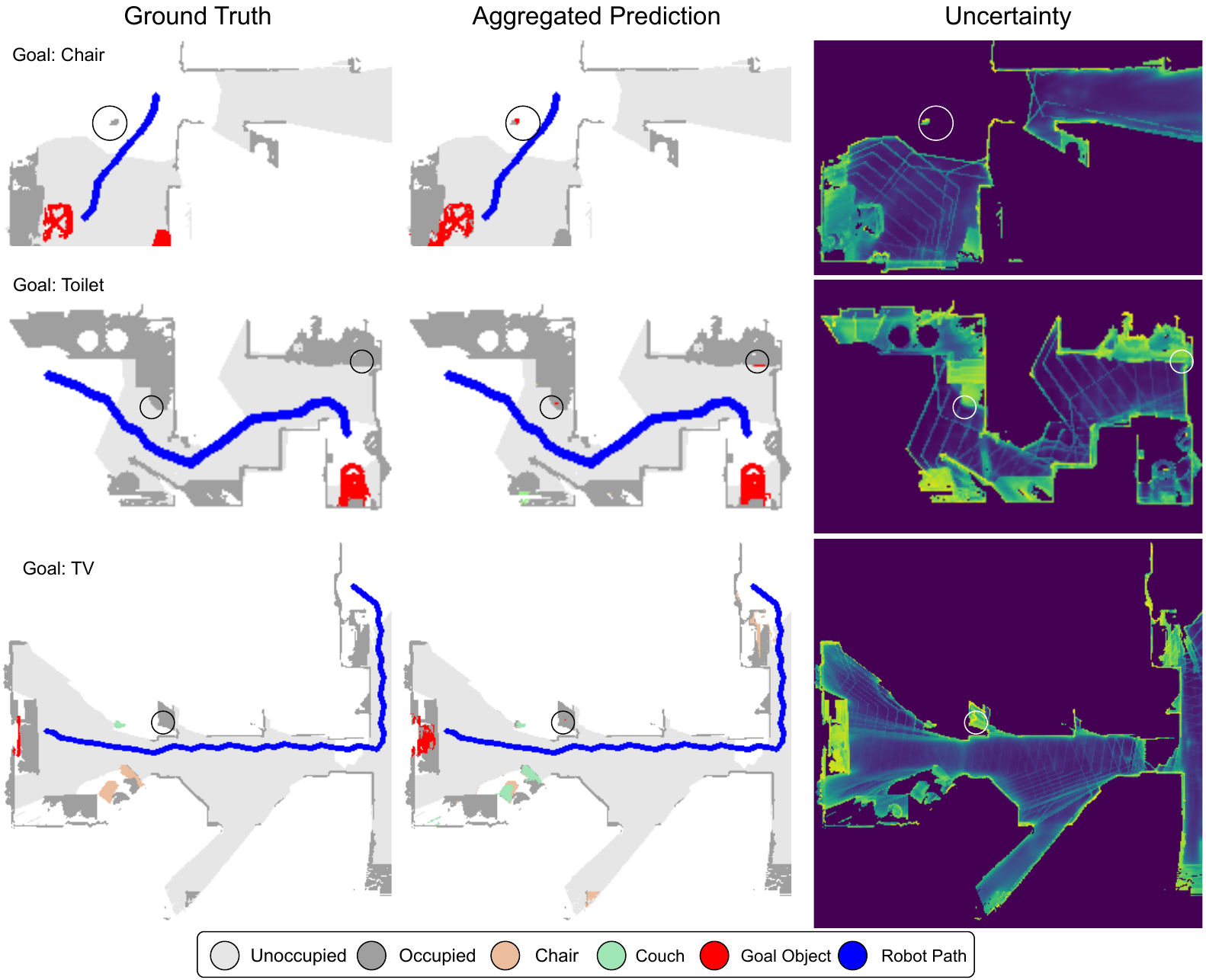}
  \caption{Semantic maps showing, from left to right, the ground truth semantics, the aggregated predictions of our Weighted Averaging approach, and the resulting uncertainty map. Circles indicate positions where a target object was falsely detected but due to the high uncertainty, no false found decision was raised. The uncertainty varies from blue-yellow corresponding to 0.0-1.0 normalized entropy.} \label{fig:map_viz}
\vspace{-0.5cm}
\end{figure}

%% file: tables/emsanet_rl.tex
\begin{table*}
\centering
\scriptsize
\caption{ObjectNav results of the RL policy trained with ground-truth semantics and deployed with EMSANet.}\label{tab:emsanet_rl}
\begin{threeparttable}
    \begin{tabularx}{\textwidth}{l|c|c|YYY|YY|Y}
        \toprule
        & \textbf{Calibrated} & \textbf{Uncertainty} &  \textbf{SR}  & \multicolumn{2}{c|}{\textbf{Found}} & \multicolumn{2}{c|}{\textbf{Detection}} & \textbf{SPL}\\
        \textbf{Aggregation} & \textbf{Probabilities} & \textbf{Found Decision} &                                                          &   FPR & FNR & \#FP & \#FN &\\
        \midrule
        \rowcolor{Gray}
        Ground Truth & \no & \no   & 75.0 & \phantom{0}3.8 & 21.2 & 0.18 & 0.00 & 27.9 \\
        Latest & \no & \no & 15.8 & 81.8 & \phantom{0}2.4 & 3.51 & 0.13 & \phantom{0}7.1 \\
        Hits/Views & \no & \no & 48.3 & 20.9 & 30.8 & 0.85 & 0.51 & 14.0 \\
        Skill Fusion~\cite{staroverov2023skill} & \no & \no & 48.4 & 33.1 & 18.5 & 0.89 & 0.46 & 16.7 \\
        Stubborn~\cite{luo2022stubborn} & \no & \no & 18.1 & 77.9 & 4.0 & 3.52 & 0.14 & \phantom{0}7.9 \\
        Latest Filtered & \yes & \no & 51.2 & 20.0 & 28.8 & 0.44 & 0.70 & 13.9 \\
        Log odds   & \yes & \yes & 52.9 & 27.1 & 20.0 & 5.26 & 0.24 & \textbf{19.2} \\
        Averaging  & \yes & \no & 28.5 & 61.2 & 10.3 & 3.36 & 0.17 & 10.7 \\
        Averaging & \no & \yes & 32.2 & 54.5 & 13.3 & 3.73 & 0.16 & 12.0 \\
        Averaging (Ours) & \yes & \yes & \underline{55.6} & 15.3 & 29.1 & 5.73 & 0.19 & \textbf{19.2} \\   
        Weighted Averaging (Ours)  & \yes & \yes & \textbf{56.4} & 15.6 & 28.0 & 6.16 & 0.18 & \textbf{19.2} \\
        \bottomrule
    \end{tabularx}
      \begin{tablenotes}[para,flushleft]
       Best and second best in bold and underlined. SR: success rate, FPR: false positive rate, NR: false positive rate, \#FP: false positives, \#FN: false negatives.
     \end{tablenotes}
 \end{threeparttable}
 \vspace{-0.5cm}
\end{table*}

%% file: 5_conclusion.tex
In this work, we identified gaps in the object search literature and showed that parts of the perception gap can be explained by the ineffective use of pretrained semantic perception models.
We proposed the incorporation of calibrated probabilities and uncertainties across map aggregation and decision making, and demonstrated their effectiveness across different perception models and policies.
The resulting methods are easy to incorporate and enable a wide family of models to further reduce the gap without additional training. As a result, they enable researchers to further close the gap. Furthermore, we made our code publicly available to aid future research.

In future work, we plan to further investigate direct policy conditioning on imperfect perception and uncertainties.
To address the overconfidence induced by training RL agents with ground truth perception, we hypothesize that providing calibrated uncertainty maps as inputs can enable the agent to make better decisions such as investigating uncertain areas more closely and to learn to make context-dependent found decisions.

%% file: 6_appendix.tex
\section{Appendix: Extended Results}\label{app:aggregation_results}

We report full results with the shortest path policy for Mask-RCNN in \tabref{tab:maskrcnn_sp} and Segformer in \tabref{tab:segformer_sp} and results using the RL Policy for both models in \tabref{tab:maskrcnn_rl} and \tabref{tab:segformer_rl}. Note that not all aggregation methods are applicable to Mask-RCNN.

\input{tables/maskrcnn_sp}
\input{tables/segformer_sp}
\input{tables/maskrcnn_rl}
\input{tables/segformer_rl}

%% file: tables/maskrcnn_sp.tex
\begin{table}[ht]
\centering
\scriptsize
\caption{ObjectNav results of the Shortest-Path Policy with Mask-RCNN.}\label{tab:maskrcnn_sp}
\begin{threeparttable}
    \begin{tabularx}{\textwidth}{l|c|c|YYY|YY}
        \toprule
        & \textbf{Calibrated} & \textbf{Uncertainty} &  \textbf{SR}  & \multicolumn{2}{c|}{\textbf{Found}} & \multicolumn{2}{c}{\textbf{Detection}}\\
        \textbf{Aggregation} & \textbf{Found Decision} & \textbf{Probabilities}                                                           &  & FPR & FNR & \#FP & \#FN \\
        \midrule
        \rowcolor{Gray}
        Ground Truth & \no & \no & 99.2    & 0.00      & 0.8      & 0.11           & 0.00           \\
        Latest & \no & \no & 46.1 & 32.2 & 21.7 & 2.11 & 0.39 \\
        Hits/Views & \no & \no & \underline{49.4} & 29.0 & 21.6 & 0.47 & 0.30  \\
        Skill Fusion~\cite{staroverov2023skill} & \no & \no & \textbf{59.2} & 13.7 & 27.1 & 0.32 & 0.55 \\
        \bottomrule
    \end{tabularx}
      \begin{tablenotes}[para,flushleft]
       Best and second best in bold and underlined. SR: success rate, FPR: false positive rate, NR: false positive rate, \#FP: false positives, \#FN: false negatives.
     \end{tablenotes}
 \end{threeparttable}
 \vspace{-0.5cm}
\end{table}

%% file: tables/segformer_sp.tex
\begin{table*}[ht]
\centering
\scriptsize
\caption{ObjectNav results of the Shortest-Path Policy with Segformer.}\label{tab:segformer_sp}
\begin{threeparttable}
    \begin{tabularx}{\textwidth}{l|c|c|YYY|YY}
        \toprule
        & \textbf{Calibrated} & \textbf{Uncertainty} &  \textbf{SR}  & \multicolumn{2}{c|}{\textbf{Found}} & \multicolumn{2}{c}{\textbf{Detection}}\\
        \textbf{Aggregation} & \textbf{Probabilities} & \textbf{Found Decision}                                                          &  & FPR & FNR & \#FP & \#FN \\
        \midrule
        \rowcolor{Gray}
        Ground Truth & \no & \no & 99.2    & 0.00      & \phantom{0}0.8      & 0.11           & 0.00           \\
        Latest & \no & \no & 41.9 & 46.8 & 11.3 & 3.04 & 0.22           \\
        Hits/Views & \no & \no & 54.6 & 11.2 & 34.2 & 0.88 & 0.16           \\
        Skill Fusion~\cite{staroverov2023skill} & \no & \no & \underline{66.0} & 11.7 & 22.3 & 0.26 & 0.47            \\
        Stubborn~\cite{luo2022stubborn} & \no & \no & 42.9  & 45.6  & 11.5 & 2.96  & 0.22  \\
        Latest Filtered & \yes & \no & 64.6 & 11.8 & 23.6 & 0.27 & 0.57           \\
        Log odds  & \yes & \yes & 63.6 & 19.9 & 16.5 & 2.18 & 0.23 \\
        Averaging  & \yes & \no &    53.1 & 34.9  & 12.0 & 2.21 & 0.23          \\
        Averaging & \no & \yes &  52.9 &  35.5 &  11.6 & 2.28 &  0.22 \\
        Averaging (Ours) & \yes & \yes & 65.8 & 6.5 & 27.7 & 2.23 & 0.23           \\
        Weighted Averaging (Ours) & \yes & \yes & \textbf{70.4} & 13.8 & 15.8 & 2.23 & 0.22 \\
        \bottomrule
    \end{tabularx}
      \begin{tablenotes}[para,flushleft]
        Best and second best in bold and underlined. SR: success rate, FPR: false positive rate, NR: false positive rate, \#FP: false positives, \#FN: false negatives.
\end{tablenotes}
 \end{threeparttable}
\vspace{-0.5cm}
\end{table*}

%% file: tables/maskrcnn_rl.tex
\begin{table}[ht]
\centering
\scriptsize
\caption{ObjectNav results of the RL policy trained with ground-truth semantics and deployed with Mask-RCNN.}\label{tab:maskrcnn_rl}
\begin{threeparttable}
    \begin{tabularx}{\textwidth}{l|c|c|YYY|YY|Y}
        \toprule
        & \textbf{Calibrated} & \textbf{Uncertainty} &  \textbf{SR}  & \multicolumn{2}{c|}{\textbf{Found}} & \multicolumn{2}{c|}{\textbf{Detection}} & \textbf{SPL} \\
        \textbf{Aggregation} & \textbf{Probabilities} & \textbf{Found Decision}                                                          &  & FPR & FNR & \#FP & \#FN & \\
        \midrule
        \rowcolor{Gray}
        Ground Truth & \no & \no  & 75.0 & \phantom{0}3.8 & 21.2 & 0.18 & 0.00 & 27.9 \\
        Latest & \no & \no & 34.7 & 38.2 & 13.6 & 2.78 & 0.49 & 11.1  \\
        Hits/Views & \no & \no & \underline{40.2} & 34.3 & 25.5 & 1.11 & 0.44 & \underline{12.2}  \\
        Skill Fusion~\cite{staroverov2023skill} & \no & \no & \textbf{46.2} & 28.8 & 25.0 & 0.47 & 0.85 & \textbf{14.5} \\
        \bottomrule
    \end{tabularx}
      \begin{tablenotes}[para,flushleft]
       Best and second best in bold and underlined. SR: success rate, FPR: false positive rate, NR: false positive rate, \#FP: false positives, \#FN: false negatives.
     \end{tablenotes}
 \end{threeparttable}
 \vspace{-0.5cm}
\end{table}

%% file: tables/segformer_rl.tex
\begin{table}[ht]
\centering
\scriptsize
\caption{ObjectNav results of the RL policy trained with ground-truth semantics and deployed with Segformer.}\label{tab:segformer_rl}
\begin{threeparttable}
    \begin{tabularx}{\textwidth}{l|c|c|YYY|YY|Y}
        \toprule
        & \textbf{Calibrated} & \textbf{Uncertainty} &  \textbf{SR}  & \multicolumn{2}{c|}{\textbf{Found}} & \multicolumn{2}{c|}{\textbf{Detection}} & \textbf{SPL} \\
        \textbf{Aggregation} & \textbf{Probabilities} & \textbf{Found Decision}                                                        &  & FPR & FNR & \#FP & \#FN \\
        \midrule
        \rowcolor{Gray}
        Ground Truth & \no & \no  & 75.0 & \phantom{0}3.8 & 21.2 & 0.18 & 0.00 & 27.9 \\
        Latest & \no & \no & 33.0 & 52.4 & 14.6 & 2.63 & 0.29 & 12.3 \\
        Hits/Views & \no & \no & 50.4 & 15.8 & 33.8 & 1.04 & 0.29 & 12.9 \\
        Skill Fusion~\cite{staroverov2023skill} & \no & \no & \textbf{54.3} & 21.3 & 24.4 & 0.45 & 0.68 & \textbf{18.1} \\
        Stubborn~\cite{luo2022stubborn} & \no & \no & 36.7 & 52.9 & 10.4 & 2.58 & 0.27 & 14.0 \\
        Latest Filtered & \yes & \no & 49.1 & 20.4 & 30.5 & 0.46 & 0.75 & 15.0 \\
        Log odds & \yes & \yes & 51.5 & 21.8 & 26.7 & 2.23 & 0.36 & \textbf{18.1} \\
        Averaging  & \yes & \no & 42.8 & 34.7 & 22.5 & 2.16 & 0.31 & 14.7 \\
        Averaging & \no & \yes &  44.9 &  33.2 & 21.9 & 2.25  & 0.31 & 16.2  \\
        Averaging (Ours) & \yes & \yes & 51.9 & 12.3 & 35.8 & 2.73 & 0.38 & 17.4 \\  
        Weighted Averaging (Ours) & \yes & \yes & \underline{52.9} & 17.6 & 29.5 & 2.44 & 0.36 & \textbf{18.1} \\
        \bottomrule
    \end{tabularx}
      \begin{tablenotes}[para,flushleft]
       Best and second best in bold and underlined. SR: success rate, FPR: false positive rate, NR: false positive rate, \#FP: false positives, \#FN: false negatives.
     \end{tablenotes}
 \end{threeparttable}
\vspace{-0.5cm}
\end{table}

%% file: main.bbl
\begin{thebibliography}{10}
\providecommand{\url}[1]{{#1}}
\providecommand{\urlprefix}{URL }
\expandafter\ifx\csname urlstyle\endcsname\relax
  \providecommand{\doi}[1]{DOI~\discretionary{}{}{}#1}\else
  \providecommand{\doi}{DOI~\discretionary{}{}{}\begingroup \urlstyle{rm}\Url}\fi

\bibitem{embodiedAIworkshop}
{CVPR Embodied AI Workshop}.
\newblock \url{https://embodied-ai.org/cvpr2023/} (2023)

\bibitem{bergstra2011algorithms}
Bergstra, J., Bardenet, R., Bengio, Y., K\'{e}gl, B.: Algorithms for hyper-parameter optimization.
\newblock In: Proc.~of the Conf.~on Neural Information Processing Systems, vol.~24 (2011)

\bibitem{chang2023goat}
Chang, M., Gervet, T., Khanna, M., Yenamandra, S., Shah, D., Min, S.Y., Shah, K., Paxton, C., Gupta, S., Batra, D., et~al.: Goat: Go to any thing.
\newblock arXiv preprint arXiv:2311.06430  (2023)

\bibitem{chaplot2020learning}
Chaplot, D.S., Gandhi, D., Gupta, S., Gupta, A., Salakhutdinov, R.: Learning to explore using active neural slam.
\newblock In: Int. Conf. on Learning Representations (2020)

\bibitem{chaplot2020object}
Chaplot, D.S., Gandhi, D.P., Gupta, A., Salakhutdinov, R.R.: Object goal navigation using goal-oriented semantic exploration.
\newblock Proc.~of the Conf.~on Neural Information Processing Systems \textbf{33}, 4247--4258 (2020)

\bibitem{chen2023not}
Chen, J., Li, G., Kumar, S., Ghanem, B., Yu, F.: How to not train your dragon: Training-free embodied object goal navigation with semantic frontiers.
\newblock Robotics: Science and Systems  (2023)

\bibitem{robothor}
Deitke, M., Han, W., Herrasti, A., Kembhavi, A., Kolve, E., et~al.: Robothor: An open simulation-to-real embodied ai platform.
\newblock In: Proc.~of the IEEE Conf.~on Computer Vision and Pattern Recognition (2020)

\bibitem{gal2016dropout}
Gal, Y., Ghahramani, Z.: Dropout as a bayesian approximation: Representing model uncertainty in deep learning.
\newblock In: Int. Conf. on Mach. Learning, pp. 1050--1059. PMLR (2016)

\bibitem{gosala2023skyeye}
Gosala, N., Petek, K., Drews-Jr, P.L., Burgard, W., Valada, A.: Skyeye: Self-supervised bird's-eye-view semantic mapping using monocular frontal view images.
\newblock In: Proc.~of the IEEE Conf.~on Computer Vision and Pattern Recognition, pp. 14,901--14,910 (2023)

\bibitem{guo2017calibration}
Guo, C., Pleiss, G., Sun, Y., Weinberger, K.Q.: On calibration of modern neural networks.
\newblock In: Int. Conf. on Mach. Learning, pp. 1321--1330. PMLR (2017)

\bibitem{he2017mask}
He, K., Gkioxari, G., Doll{\'a}r, P., Girshick, R.: Mask r-cnn.
\newblock In: Int.~Conf.~on Computer Vision, pp. 2961--2969 (2017)

\bibitem{honerkamp2024language}
Honerkamp, D., Buchner, M., Despinoy, F., Welschehold, T., Valada, A.: Language-grounded dynamic scene graphs for interactive object search with mobile manipulation.
\newblock {IEEE} Robotics and Automation Letters  (2024)

\bibitem{hurtado2022semantic}
Hurtado, J.V., Valada, A.: Semantic scene segmentation for robotics.
\newblock In: Deep learning for robot perception and cognition, pp. 279--311. Elsevier (2022)

\bibitem{jiang2018rednet}
Jiang, J., Zheng, L., Luo, F., Zhang, Z.: Rednet: Residual encoder-decoder network for indoor rgb-d semantic segmentation.
\newblock arXiv preprint arXiv:1806.01054  (2018)

\bibitem{kolve2017ai2}
Kolve, E., Mottaghi, R., Han, W., VanderBilt, E., Weihs, L., Herrasti, A., Deitke, M., Ehsani, K., Gordon, D., Zhu, Y., et~al.: Ai2-thor: An interactive 3d environment for visual ai.
\newblock arXiv preprint arXiv:1712.05474  (2017)

\bibitem{lakshminarayanan2017simple}
Lakshminarayanan, B., Pritzel, A., Blundell, C.: Simple and scalable predictive uncertainty estimation using deep ensembles.
\newblock Proc.~of the Conf.~on Neural Information Processing Systems \textbf{30} (2017)

\bibitem{luo2022stubborn}
Luo, H., Yue, A., Hong, Z.W., Agrawal, P.: Stubborn: A strong baseline for indoor object navigation.
\newblock In: Int.~Conf.~on Intelligent Robots and Systems, pp. 3287--3293. IEEE (2022)

\bibitem{maksymets2021thda}
Maksymets, O., Cartillier, V., Gokaslan, A., Wijmans, E., Galuba, W., Lee, S., Batra, D.: Thda: Treasure hunt data augmentation for semantic navigation.
\newblock In: Proc.~of the IEEE Conf.~on Computer Vision and Pattern Recognition, pp. 15,374--15,383 (2021)

\bibitem{mccormac2017semanticfusion}
McCormac, J., Handa, A., Davison, A., Leutenegger, S.: Semanticfusion: Dense 3d semantic mapping with convolutional neural networks.
\newblock In: Int.~Conf.~on Robotics \& Automation, pp. 4628--4635 (2017)

\bibitem{naeini2015obtaining}
Naeini, M.P., Cooper, G., Hauskrecht, M.: Obtaining well calibrated probabilities using bayesian binning.
\newblock Proc.~of the National Conference on Artificial Intelligence \textbf{29}(1) (2015)

\bibitem{nallapareddy2023evcenternet}
Nallapareddy, M.R., Sirohi, K., Drews, P.L., Burgard, W., Cheng, C.H., Valada, A.: Evcenternet: Uncertainty estimation for object detection using evidential learning.
\newblock In: Int.~Conf.~on Intelligent Robots and Systems, pp. 5699--5706. IEEE (2023)

\bibitem{ramakrishnan2022poni}
Ramakrishnan, S.K., Chaplot, D.S., Al-Halah, Z., Malik, J., Grauman, K.: Poni: Potential functions for objectgoal navigation with interaction-free learning.
\newblock In: Proc.~of the IEEE Conf.~on Computer Vision and Pattern Recognition. IEEE (2022)

\bibitem{ramakrishnan2021hm3d}
Ramakrishnan, S.K., Gokaslan, A., Wijmans, E., Maksymets, O., et~al.: Habitat-matterport 3d dataset ({HM}3d): 1000 large-scale 3d environments for embodied {AI}.
\newblock In: Proc.~of the Conf.~on Neural Information Processing Systems (2021)

\bibitem{ramrakhya2023pirlnav}
Ramrakhya, R., Batra, D., Wijmans, E., Das, A.: Pirlnav: Pretraining with imitation and rl finetuning for objectnav.
\newblock In: Proc.~of the IEEE Conf.~on Computer Vision and Pattern Recognition, pp. 17,896--17,906 (2023)

\bibitem{raychaudhuri2024mopa}
Raychaudhuri, S., Campari, T., Jain, U., Savva, M., Chang, A.X.: Mopa: Modular object navigation with pointgoal agents.
\newblock In: Proc. of the IEEE/CVF Winter Conference on Applications of Computer Vision, pp. 5763--5773 (2024)

\bibitem{roberts2021hypersim}
Roberts, M., Ramapuram, J., Ranjan, A., Kumar, A., Bautista, M.A., Paczan, N., Webb, R., Susskind, J.M.: Hypersim: A photorealistic synthetic dataset for holistic indoor scene understanding.
\newblock In: Proc.~of the IEEE Conf.~on Computer Vision and Pattern Recognition, pp. 10,912--10,922 (2021)

\bibitem{fabian22exploration}
Schmalstieg, F., Honerkamp, D., Welschehold, T., Valada, A.: Learning long-horizon robot exploration strategies for multi-object search in continuous action spaces.
\newblock Proceedings of the International Symposium on Robotics Research (ISRR)  (2022)

\bibitem{schmalstieg2023learning}
Schmalstieg, F., Honerkamp, D., Welschehold, T., Valada, A.: Learning hierarchical interactive multi-object search for mobile manipulation.
\newblock {IEEE} Robotics and Automation Letters  (2023)

\bibitem{seichter2022efficient}
Seichter, D., Fischedick, S.B., K{\"o}hler, M., Gro{\ss}, H.M.: Efficient multi-task rgb-d scene analysis for indoor environments.
\newblock In: 2022 Int. joint conference on neural networks (IJCNN), pp. 1--10. IEEE (2022)

\bibitem{sensoy2018evidential}
Sensoy, M., Kaplan, L., Kandemir, M.: Evidential deep learning to quantify classification uncertainty.
\newblock Proc.~of the Conf.~on Neural Information Processing Systems \textbf{31} (2018)

\bibitem{sirohi2024uplam}
Sirohi, K., B{\"u}scher, D., Burgard, W.: uplam: Robust panoptic localization and mapping leveraging perception uncertainties.
\newblock arXiv preprint arXiv:2402.05840  (2024)

\bibitem{sirohi2023uncertainty2}
Sirohi, K., Marvi, S., B{\"u}scher, D., Burgard, W.: Uncertainty-aware lidar panoptic segmentation.
\newblock In: Int.~Conf.~on Robotics \& Automation, pp. 8277--8283. IEEE (2023)

\bibitem{sirohi2023uncertainty}
Sirohi, K., Marvi, S., B{\"u}scher, D., Burgard, W.: Uncertainty-aware panoptic segmentation.
\newblock {IEEE} Robotics and Automation Letters \textbf{8}(5), 2629--2636 (2023)

\bibitem{song2015sun}
Song, S., Lichtenberg, S.P., Xiao, J.: Sun rgb-d: A rgb-d scene understanding benchmark suite.
\newblock In: Proc.~of the IEEE Conf.~on Computer Vision and Pattern Recognition, pp. 567--576 (2015)

\bibitem{staroverov2023skill}
Staroverov, A., Muravyev, K., Yakovlev, K., Panov, A.I.: Skill fusion in hybrid robotic framework for visual object goal navigation.
\newblock Robotics \textbf{12}(4), 104 (2023)

\bibitem{szot2021habitat}
Szot, A., Clegg, A., Undersander, E., Wijmans, E., et~al.: Habitat 2.0: Training home assistants to rearrange their habitat.
\newblock arXiv preprint arXiv:2106.14405  (2021)

\bibitem{thrun2002probabilistic}
Thrun, S., Burgard, W.: Probabilistic robotics.
\newblock Communications of the ACM \textbf{45}(3), 52--57 (2002)

\bibitem{wani2020multion}
Wani, S., Patel, S., Jain, U., Chang, A.X., Savva, M.: Multi-on: Benchmarking semantic map memory using multi-object navigation.
\newblock In: Proc.~of the Conf.~on Neural Information Processing Systems (2020)

\bibitem{xie2021segformer}
Xie, E., Wang, W., Yu, Z., Anandkumar, A., Alvarez, J.M., Luo, P.: Segformer: Simple and efficient design for semantic segmentation with transformers.
\newblock Proc.~of the Conf.~on Neural Information Processing Systems \textbf{34}, 12,077--12,090 (2021)

\bibitem{habitatchallenge2023}
Yadav, K., Krantz, J., Ramrakhya, R., Ramakrishnan, S.K., Yang, J., et~al.: Habitat challenge 2023.
\newblock \url{https://aihabitat.org/challenge/2023/} (2023)

\bibitem{yamauchi1997frontier}
Yamauchi, B.: A frontier-based approach for autonomous exploration.
\newblock In: Proc.~of the IEEE Int.~Symp. on Comput. Intell. in Rob. and Aut. (1997)

\bibitem{ye2021auxiliary}
Ye, J., Batra, D., Das, A., Wijmans, E.: Auxiliary tasks and exploration enable objectgoal navigation.
\newblock In: Int.~Conf.~on Computer Vision, pp. 16,117--16,126 (2021)

\bibitem{younes2023catch}
Younes, A., Honerkamp, D., Welschehold, T., Valada, A.: Catch me if you hear me: Audio-visual navigation in complex unmapped environments with moving sounds.
\newblock {IEEE} Robotics and Automation Letters \textbf{8}(2), 928--935 (2023)

\bibitem{zemskova2024interactive}
Zemskova, T., Staroverov, A., Muravyev, K., Yudin, D., Panov, A.: Interactive semantic map representation for skill-based visual object navigation.
\newblock IEEE Access  (2024)

\bibitem{Zhou_2017_CVPR}
Zhou, B., Zhao, H., Puig, X., Fidler, S., Barriuso, A., Torralba, A.: Scene parsing through ade20k dataset.
\newblock In: Proc.~of the IEEE Conf.~on Computer Vision and Pattern Recognition (2017)

\bibitem{zhou2023esc}
Zhou, K., Zheng, K., Pryor, C., Shen, Y., Jin, H., Getoor, L., Wang, X.E.: Esc: Exploration with soft commonsense constraints for zero-shot object navigation.
\newblock arXiv preprint arXiv:2301.13166  (2023)

\bibitem{zhu2022navigating}
Zhu, M., Zhao, B., Kong, T.: Navigating to objects in unseen environments by distance prediction.
\newblock In: Int.~Conf.~on Intelligent Robots and Systems, pp. 10,571--10,578. IEEE (2022)

\bibitem{zurn2022trackletmapper}
Z{\"u}rn, J., Weber, S., Burgard, W.: Trackletmapper: Ground surface segmentation and mapping from traffic participant trajectories.
\newblock In: Proc.~of the Conference on Robot Learning (2022)

\end{thebibliography}
